\documentclass[11pt]{article}
\usepackage{coling2020}
\usepackage{times}
\usepackage{url}
\usepackage{latexsym}
\usepackage{multirow}
\usepackage{amsfonts}
\usepackage{graphicx}
\usepackage{bm}
\makeatletter 
  \newcommand\figcaption{\def\@captype{figure}\caption} 
  \newcommand\tabcaption{\def\@captype{table}\caption} 
\makeatother

\title{A Heterogeneous Graph with Factual, Temporal and Logical Knowledge for Question Answering Over Dynamic Contexts
}
\author{Wanjun Zhong$^1$\thanks{\ \ \ Work done while this author was an intern at Microsoft Research.} , Duyu Tang$^2$, Nan Duan$^2$, Ming Zhou$^2$\\
	\bf Jiahai Wang$^1$ and Jian Yin$^1$\\
	$^1$ The School of Data and Computer Science, Sun Yat-sen University.\\
	Guangdong Key Laboratory of Big Data Analysis and Processing, Guangzhou, P.R.China\\
	$^2$ Microsoft Research \\
	{\tt \{zhongwj25@mail2,wangjiah@mail,issjyin@mail\}.sysu.edu.cn}\\
	{\tt \{dutang,nanduan,mingzhou\}@microsoft.com}}

\date{}

\begin{document}
\maketitle
\begin{abstract}
We study question answering over a dynamic textual environment. 
Although neural network models achieve impressive accuracy via learning from input-output examples, they rarely leverage various types of knowledge and are generally not interpretable.
In this work, we propose a graph-based approach, where a heterogeneous graph is automatically built with factual knowledge of the context, temporal knowledge of the past states, and logical knowledge that combines human-curated knowledge bases and rule bases.
We develop a graph neural network over the constructed graph, and train the model in an end-to-end manner.
Experimental results on a benchmark dataset show that the injection of various types of knowledge improves a strong neural network baseline.
An additional benefit of our approach is that the graph itself naturally serves as a rational behind the decision making.
\end{abstract}

\section{Introduction}
In this work, we study the problem of 
question answering
over a dynamic textual environment, where the state of participants and their relationships in the environment evolve through time.
The problem is a good testbed to measure ability of natural language understanding systems in reasoning about casual effects implicitly expressed in text, and is also important for tasks like effect prediction and procedure execution and evaluation.
Reasoning over a dynamic world is challenging because it requires a model to not only understand past states of participants in the world and the effect of new evidence, but also avoid the state of change which conflicts with the universal law of the world (such as an entity cannot be located at somewhere before it is created). 

Existing approaches 
are dominated by neural network based approaches \cite{henaffWSBL16,seo2016query,bosselut2018,dalviHTYC18,tandonDGYBC18,das2019kgmrc}, where sequential and attentional architectures are used to model the interaction between the question and the context.
Neural models can be conventionally trained in an end-to-end manner, and prove to generalize well empirically with remarkably high accuracy after trained on input-output examples.
However, they are commonly accused of lack of the ability to leverage human knowledge and are generally not interpretable.
On the other hand, expert systems with  knowledge bases and logical rules are interpretable and do not rely on training data \cite{russell2002artificial}, however, they cannot generalize beyond what is manually defined in rules.
This motivates us to take the best of both worlds.

%

In this paper, we present an approach that 
injects knowledge and rules in a neural model for question answering.
The key idea is to use graph neural network as the pivot, where the graph is automatically constructed by using various types of knowledge, and representation learning over the graph is implemented with neural network. 
Specifically, a graph conveys three kinds of knowledge, including factual knowledge about the context which is obtained by information extraction, temporal knowledge about the past states of participants through transformation in semantic vector space, and logical knowledge which is obtained by the inference outcomes of a generic rule base grounded on an external knowledge base like VerbNet \cite{schuler2005verbnet}.
Representations of nodes on the graph are initially calculated with BERT \cite{devlin2018bert}, and message passing and aggregation are implemented with graph neural network \cite{scarselliGTHM09}.


We conduct experiments on PROPARA \cite{dalviHTYC18},
a benchmark dataset for reasoning the state of entities in procedural text. 
We develop our system based on ProGlobal \cite{dalviHTYC18}, a strong neural network method which is open-sourced. 
We show that our approach, which leverages graph networks and various types of knowledge, achieves $61.0\%$ F1 score, which improves ProGlobal by xxx.
We make an ablation study and 
observe that the integration of the various types of knowledge brings significant improvements.

\section{Task Formulation \& Dataset}
\label{section:task}

\begin{figure}[htb] 
  \begin{minipage}[b]{0.5\textwidth} 
    \centering 
	\includegraphics[width=1.0\textwidth]{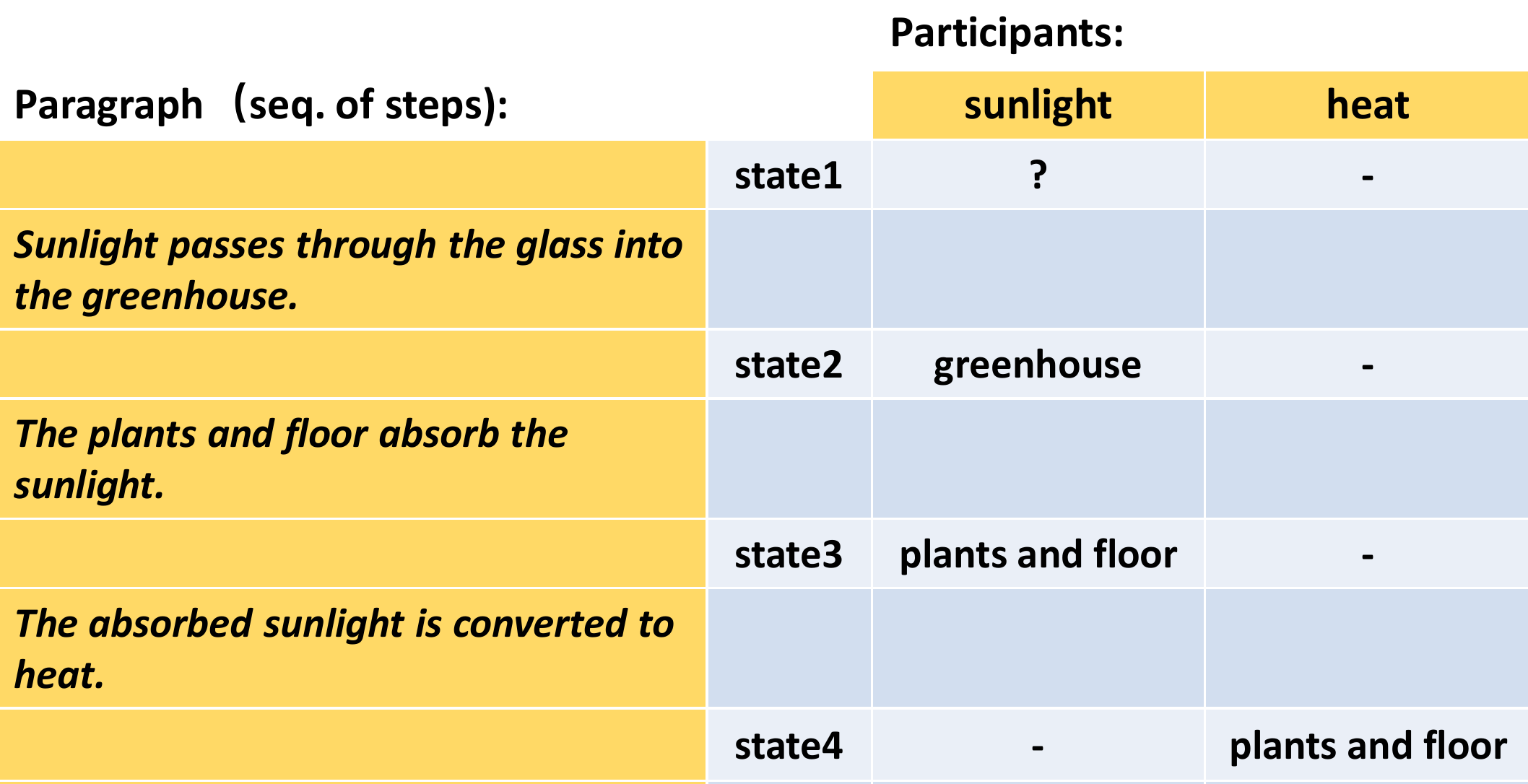}
	\caption{An example from PROPARA dataset.  }
	\label{fig:dataset-example}

  \end{minipage}%
  \begin{minipage}[b]{0.5\textwidth} 
    \centering

	\begin{tabular}{l|r}
		\hline
		Total Annotations & 81,345 \\
		Total paragraph & 488\\
		Paragraph (Train/ Dev/Test) & (391/43/54)\\
		Domains & 183 \\
		Total sentences &  3,300\\
		Avg sentences per paragraph &6.7\\
		Avg entities per paragraph& 4.17 \\
		\hline        
	\end{tabular}
	\tabcaption{Statistics of PROPARA dataset.}
	\label{table:data-statistics}

  \end{minipage} 
\end{figure}

We study on the PROPARA \cite{dalviHTYC18} dataset. 
The task is to infer the state of a specific participant given procedural text. An example is given in Figure \ref{fig:dataset-example}. 
Each grid in the row describes the location and existence of a given participant before and after each time step (``-" means ``no exist" and ``?" means ``unknown"). For instance, sunlight is located at greenhouse after time step 1 (first sentence).

	
PROPARA \cite{dalviHTYC18} is built based on natural procedural text. Crowd workers annotated the location of given participants at each time step (sentence) in the paragraph. 
Table \ref{table:data-statistics} summarizes the statistics of the PROPARA dataset.
The evaluation metrics proposed by 
 \newcite{mishra2018tracking} and \newcite{tandonDGYBC18} are equivalent to predict the state and location of participants via answering the following seven questions. 

$\bullet$ \textbf{Q1:} \textit{Which participants are the inputs of the overall process?}
	
	$\bullet$ \textbf{Q2:} \textit{Which participants are the outputs of the overall process?}

	$\bullet$ \textbf{Q3:} \textit{When and where does the conversion occur? }

	$\bullet$ \textbf{Q4:} \textit{When and where does the movement occur?}
	
	$\bullet$ \textbf{Q5:}  \textit{Is the participant ever created (destroyed, moved) in the overall process?
		}
		
	$\bullet$ \textbf{Q6:} \textit{In which time step is the participant created (moved, destroyed)?}
	
	$\bullet$ \textbf{Q7:} \textit{Where is the participant created (destroyed at, moved from/to)? }


\section{Related Work}
Approaches in literature are dominated by neural models.
Before PROPARA was constructed, representative neural models update the representation of state of participant with the action, and also model historical information of entities with recurrent attention. We describe the following representative works. 
EntNet \cite{henaffWSBL16} maintains and updates a representation of the state of the entity with a gating mechanism as the model reading text. 
QRN \cite{seo2016query} updates the representation of the question as the model reading a passage.
NPN \cite{bosselut2018} learns explicit action representations as functional operators.

Along with releasing PROPARA, 
\newcite{dalviHTYC18} introduce two neural models, ProLocal and ProGlobal.
ProLocal takes a sentence and a participant as the input, and predicts state change type and location spans.
The model outputs are post-processed with persistence/inertia rules. 
Compared to ProLocal, ProGlobal further encodes the entire paragraph and the location of each word, and predicates state (``not exist'', ``unknown location'', ``known location'').
\newcite{tandonDGYBC18} develop ProStruct, which predicts state change (including ``MOVE'', ``CREATE'', ``DESTROY'', ``NONE'') and detects the locations of beginning and ending words with commonsense-based constraints.
 \newcite{gupta2019tracking}
 track the state of each participant and uses a neural CRF (Conditional Random Field) to model the global information of participant changes explicitly.
 KG-MRC \cite{das2019kgmrc} learns the state of each participant by querying a machine reading comprehension model and keeping track of entities over a knowledge graph. 
Our work differs from KG-MRC in that our graph has more abundant types of knowledge including explicit factual edges, temporal edges, and logical edges.

A representative logic-based model on PROPARA is ProComp \cite{clark2018}, which is largely neglected in the literature.
ProComp processes paragraph with OpenIE and SRL, and derives rule bases from VerbNet with manual checking and correcting. It performs reasoning with four commonsense laws and three pragmatics of discourse. 
Our graph construction process is largely inspired by their great work.
%
%


\section{Approach}


Following the dominant workflow in the literature, we solve the task by predicting the state change and location span of a participant. 
The former consists of four types of state change (\textit{CREATE}, \textit{DESTROY}, \textit{MOVE}, \textit{NONE}), which indicate how the state of participants influenced by events happened on each time step (sentence).  The latter indicates the location of participants before and after each time step.

\begin{figure}[h]
	\centering
	\includegraphics[width=\linewidth]{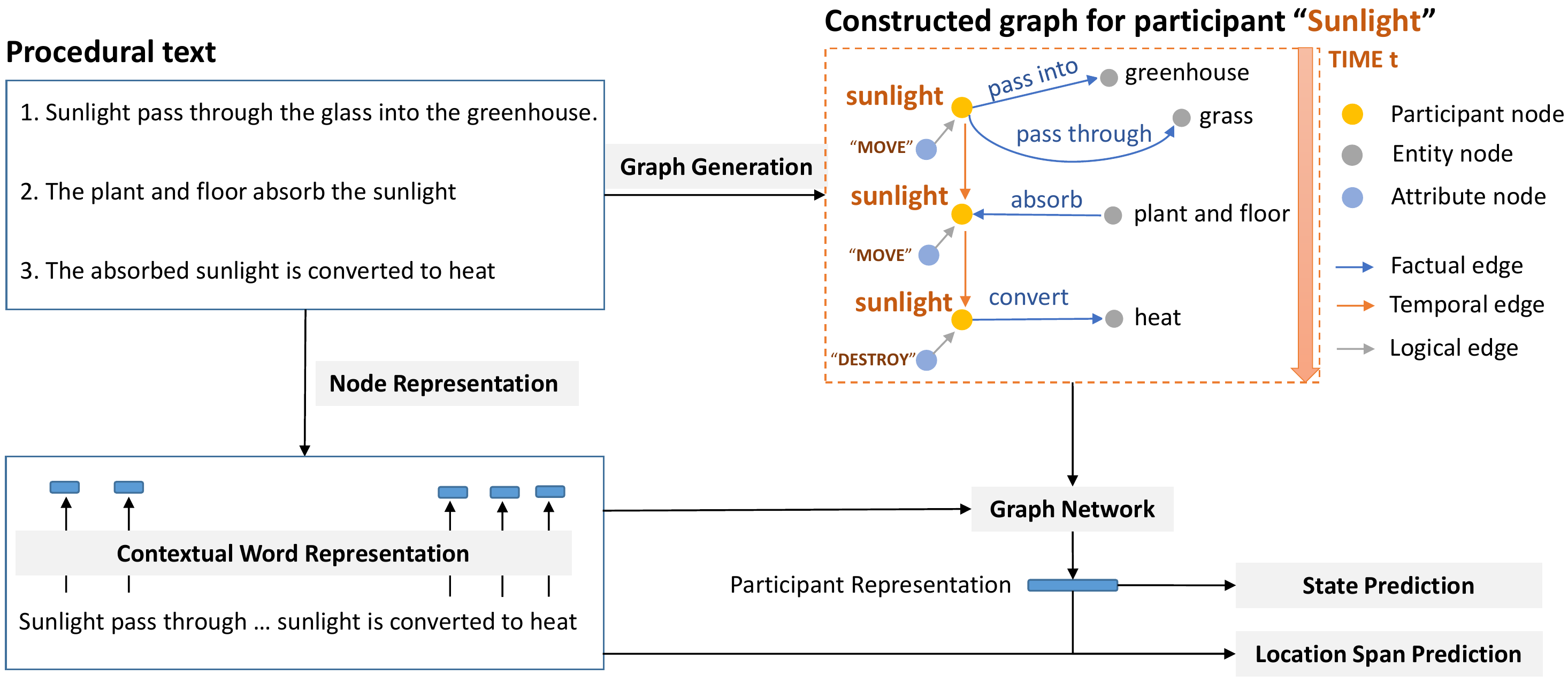}
	\caption{An overview of the pipeline of our approach.
	 	Example of graph constructed with given paragraph. The orange node indicate the specific participant. The orange line indicate \textit{temporal edge} and blue line indicate \textit{verb edge}. The grey line indicate \textit{attribute edge}. Each participant is given as a prior. \textit{Participant nodes} are related to several \textit{entity nodes} extracted from procedural text at each time step and are related to \textit{attribute nodes} deduced from symbolic system.}
	\label{fig:overview}
\end{figure}

At a high level, the design of our approach contains three main components: \textit{graph construction}, \textit{representation learning over the graph} and \textit{prediction}, as given in Figure \ref{fig:overview}. 
Given a procedural paragraph and a participant as input, the graph construction component constructs a participant-specific graph.
After constructing the graph, we represent each node with contextual word representations and calculate the representations of participants with a graph neural network.
With learned representations of participants, we have two prediction models to predict the state and location of them at each time step. 

%


\subsection{Graph Construction}
\label{sec:graph-construction-system}


The upper right part of Figure \ref{fig:overview} shows an example of a constructed graph for the participant ``\textit{sunlight}'' given a procedural text on the left as context. 
The constructed graph not only relates the participants to other entities involved in events at each time step, but also persists the temporal consistency between participant nodes at different time steps. The graph is constructed in a participant-oriented manner with three types of edges, namely factual edges, temporal edges, and logical edges, which we will detail later.  

\paragraph{Notation}
A graph contains six main components $G = (P,S,A,E_{ps},E_{pp^{'}},E_{pa})$, where 
$P=\{p^{t}\}_{t=1:|T|}$ denotes \textit{participant nodes} at all time steps, 
$t$ denotes current time step (sentence), 
$S=\{s^{j}\}_{j=1:|S|}$ denotes a set of \textit{entity nodes} (noun or phrase)
that are related to participants connected by ``\textit{factual edge}" $E_{ps}$, 
$A$ denotes a set of ``\textit{attribute nodes}" $a$, which relates to participants $P $ with \textit{``logical edge"} $E_{pa}$, and 
$S_{p^{t}}$ denotes the subset of $S$ that have all entities related to $p^{t}$. 
We further define ``\textit{temporal edge}" $E_{pp^{'}}$ to indicate a set of edges that persists temporal continuity between $p^{t-1}$ and $p^{t}$. 

\paragraph{Factual Edges}
For each sentence that mentions the target participant, entity nodes and their factual edges connected to participants are automatically extracted via by OpenIE \cite{Stanovsky2018SupervisedOI} or SRL (Semantic Role Labeling) \cite{shi2019simple} toolkits.
To increase the coverage of the graph, we make an extension by using POS tagger and dependency relations \cite{manning2014stanford} to construct the related tuples if they cannot be found by OpenIE or SRL.
To filter tuples related to the current participant, we apply a soft-match mechanism to align the participant and inferred entities.
In this way, we obtain a set of {entity nodes} $S$ (noun or phase) related to current participant with {factual edges} $E_{ps}$. 

\paragraph{Temporal Edges} 
To model the temporal relationship between the same participant at different time steps, we define {temporal edge} $E_{pp^{'}}$ as a set of edges that connect participants at time step $t-1$ to participants at time step $t$. 

\begin{figure}[h]
	\centering
	\includegraphics[width=0.5\linewidth]{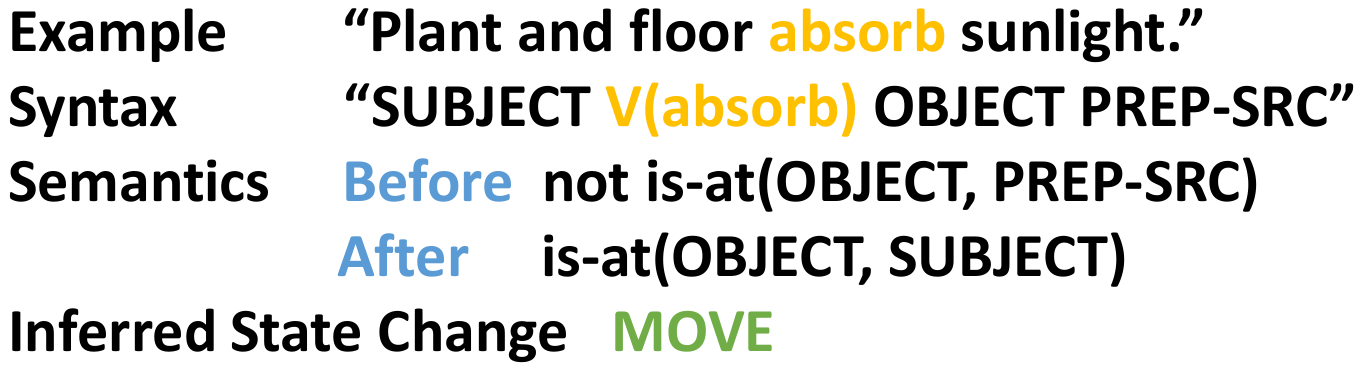}
	\caption{Example of Semantic Lexicon. The example describe an event happens on participant ``sunlight". The syntax of this sentence is extracted via SRL toolkits. Semantics of this syntax is defined in Semantic Lexicon. With extracted syntax and corresponding semantics, we can infer that the state change of participant ``sunlight" is ``MOVE". }
	\label{fig:example-semantic-lexicon}
\end{figure}

\paragraph{Logical Edges}
We largely follow \newcite{clark2018} and use Semantic Lexicon to infer attribute nodes, which include the state of the participants at different time steps. Semantic Lexicon is an expert-curated rule base built based on VerbNet, describing the state of participants changed by different actions. 
In Semantic Lexicon, each query consists of a verb $V$ and a syntax pattern $(S,V,O,(PREP\ NP)*)$, describing the core action and the syntactic elements in a sentence. With a matched query, the Semantic Lexicon states the before and after state of an entity.
Figure \ref{fig:example-semantic-lexicon} shows an example of query and statement of Semantic Lexicon. 
With a verb and a pattern described by each extracted tuple from SRL or OpenIE, our system will query the Semantic Lexicon for the state and state change of participant.
In this way, we construct a set of {attribute nodes} $A$ with inferred state change.

\paragraph{Graph Complement}
Sometimes the state change is not explicitly mentioned, but the location can be inferred from the text. In order to complete the graph, we define several commonsense rules to help the system to deduce states by the locations reversely. The commonsense rules are described as follows:

	$\bullet$ 
	\textbf{R1}: The inferred state change is \textit{NONE} if the before and after locations are the same.
	
	$\bullet$  \textbf{R2}: The inferred state change is \textit{MOVE} if the before and after locations exist and are different.
	
	$\bullet$  \textbf{R3}: The inferred state change is \textit{CREATE} if the participant does not exists before but has after location inferred from the current step.

\subsection{Graph-Based Presentation Learning}
\label{section:graph-representation}
We describe how we learn the representation of participants based on the constructed graph. 
\paragraph{Background: Graph Network}
We first introduce the common notations in graph network \cite{gori2005new,scarselliGTHM09,li2015gated,battaglia2018,song2018exploring,de2018question}, which will be applied in the following components.

Graph network framework represents the reasoning framework built based on the relational graph structure. 
We partly follow the notations mentioned by 
Battaglia et al. \cite{battaglia2018}
 and take the ``graph" as a directed and attributed multi-graph. 
Thus, the graph is denoted as $G=(V,E)$, where $V= \{v_i\}_{i=1:N^v}$ represents a set of nodes and $E= \{e_k\}_{k=1:N^e}$ denotes a set of edges. 
Each $a_{v_i}$ represents the attribute of a specific node. The attribute of a given node or edge may have multi-dimensional attributes. 
$NBR(v_i)$ here denotes the set of neighbors that have incoming or outgoing edges to node $i$. $CO(v_i)$ denotes all the incoming and outgoing edges from $v_i$. 
As described in
Battaglia et al. \cite{battaglia2018} 
, the node representation is learned following the below recurrence equation for time step $t$, where $l_v$ and $l_{CO(v)}$ denotes the label for node $v$ and $CO(v)$ respectively. 
\begin{equation}
h^{(t)}_v = f^*(l_v,l_{CO(v)},l_{NBR(v)},h^{(t-1)}_{NBR(v)})
\end{equation}
where $f^*()$ denoted update function aggregating information.

\paragraph{Node Representations} 
There are three types of nodes in our graph, namely ``\textit{participant nodes}", ``\textit{entity nodes}", and ``\textit{attribute nodes}". The first two types of nodes are text spans occurred in the paragraph, where the former is given as a prior and the latter is extracted by OpenIE or SRL toolkits. We leverage the contextual word representations to initialize the representation of them.
Since attributes come from a fixed vocabulary, we adopt an additional word embedding matrix to encode them.
\par Specifically, in order to obtain the representation of a word span, we first derive the contextual representation of the paragraph. We first represent each word as word embedding $v_w$. For $v_w$, we apply the concatenation of Glove embedding and contextual word embedding from BERT. 
Similar to
Mishra et al. \cite{mishra2018tracking}
, we further concatenate the representation with position embedding $v_d$ and sentence indicator embedding $v_s$. 
$v_d$ describes the relative distance of each word to the participant and $v_s$ indicates the position of the current sentence in the paragraph.  
Afterwards, at time step $t$, the contextual word embeddings $\widetilde{H}^D_t$ of all words in a procedural text are obtained via BiLSTM, which is defined as:
\begin{equation}
\widetilde{H}^D_t = [\widetilde{h}^1_t,\widetilde{h}^2_t..., \widetilde{h}^{|D|}_t]
\label{con:context-rep}
\end{equation}
where $|D|$ denotes number of words in document $D$.
\paragraph{Representation Learning over Graph}
In this part, we introduce how the representation of a given participant is learned via graph network. 

We denote $p^{t}$ as the participant $p$ at the $t$-th time step.
The neighbors of participant $p^{t}$
can be defined below according to the definition as mentioned in the previous subsection.
\begin{equation}
NBR(p^t) = \{p^{t-1}\} \cup S_{p^t} \cup A_{p^t}\label{con:nbr}
\end{equation}
where $p^t$ denotes participant node at time step $t$. $S_{p^t}$ and $A_{p^t}$ denote neighboring entity nodes and attribute nodes of $p^t$, respectively. 
\begin{equation}
CO(p^t) = E_{ps}(p^t) \cup E_{pp^{'}}(p^t) \cup E_{pa}(p^t)\label{con:co}
\end{equation}
where $E_{ps}(p^t)$, $E_{pp^{'}}(p^t)$ and $E_{pa}(p^t)$ denote the verb edges, the temporal edges and the attribute edges, respectively.

The representation of participant nodes are learned by the aggregation of information from neighbors at each time step. 
We develop two mechanisms for the representation learning: (1) recurrent unit to model information aggregated by \textit{``temporal edge"} $E_{pp^{'}}$ and \textit{``verb edge"} $E_{ps}$; (2) relational attention mechanism which acts as a fusion block to 
integrate attribute information from $A$ and the participant information $H^{'}_{p^t}$ at different time steps.
\par
Firstly, $H^{'}_{p^t}$ is calculated by accumulating the information from the previous time step (i.e. $H^{'}_{p^{t-1}}$) and the information propagated by a verb edge at the current time step (i.e., $H_{ps}$). 
$H_{ps}$ is the concatenation of three components, including the contextual representation of participants, entities and verbs. 
These representations are calculated by the sum of contextual representation of participant words, entity words and verb words, respectively.
For each of them, the contextual representation
is calculated by the sum of the contextual representations at their corresponding positions in the text.
Afterwards, the calculation of the recurrent unit is given as below, where $U^z$, $W^z$, $U^r$, $W^r$, $U^h$ and $W^h$ are model parameters.
\begin{equation}
\left\{
\begin{array}{lr}
z_t = \sigma(H_{ps}U^z+H^{'}_{p^{t-1}}W^z) &\\
r_t = \sigma(H_{ps}U^r+H^{'}_{p^{t-1}}W^r) \\
\widetilde{H}^{'}_{p^{t}} = tanh(H_{ps}U^h+(r_t*H^{'}_{p^{t-1}})W^h \\
H^{'}_{p^{t}}=(1-z_t)*H^{'}_{p^{t-1}}+z_t*\widetilde{H}^{'}_{p^{t}} & 
\end{array}
\right.
\end{equation}

\par Secondly,  to integrate the attribute inferred from the graph construction system into the representation of participants, we apply a graph relational attention mechanism \cite{velivckovic2017graph}.
In our graph, an attribute node represents the state change of a participant. 
Thus, we regard an attribute as the operation for modifying the meaning of the participants at previous time steps \cite{mitchell2010composition}.
Specifically, we take the attribute embedding $h^a$ for $a \in A$ as the query, and take the representation of participants $[H^{'}_{p^0},H^{'}_{p^1},..H^{'}_{p^{t-1}}]$ at previous time steps as the memory.
We calculate $H^A_{p^t}$ using weighted sum over the memory $H^A_{p^t} = \sum_{i = 0}^{t-1}\alpha_{i}H^{'}_{p^i}$, where the weight of $i$-th memory cell $\alpha_{i}$ is calculated by a dot-product function with the key. Then it is followed by a linear and an activation function.
The final representation of participant at time step $t$ is the concatenation of $H^{'}_{p^t}$ and $H^A_{p^t}$.
%



\subsection{Prediction Model}
\label{section:prediction}
We have two prediction models, designed for the prediction of state change and state (location and existence), respectively. The prediction model takes the contextual representation and representation of participant learned from graph network as input, and outputs the state change and location span of the participant at current time step.

\paragraph{State Change Predictor}
We take the advantages of ProGlobal \cite{tandonDGYBC18} and ProStruct \cite{mishra2018tracking}. 
We adopt a multi-task learning objectives consisting of two main predictors. (1) state change predictor classifies the state change from one of four classes: \textit{MOVE}, \textit{CREATE}, \textit{DESTROY}, \textit{NONE}. (2) location state predictor predicts state of location from one of three classes: \textit{not exist}, \textit{location unknown}, \textit{location known}. 

These two predictors take the representation $H_{p^{t}}$ and the predicted category probability vector $p^1_{t-1}\in \mathbb{R}^4$ or $p^2_{t-1}\in \mathbb{R}^3$ from last time step as inputs. We apply node-wise classification over the inputs, where $p^1_t$ and $p^2_t$ denote the category probability vectors for state change and state of location separately. These probabilities are calculated as follows, where $W_1$ and $W_2$ are model parameters and $b_1$ and $b_2$ are bias vectors.
\begin{equation}
p^1_t = softmax(W_{1} \cdot [H_{p^{t}};p^1_{t-1}] +b_{1})
\label{con:sc-predictor}
\end{equation}
\begin{equation}
p^2_t = softmax(W_{2} \cdot [H_{p^{t}};p^2_{t-1}] +b_{2})
\label{con:sl-predictor}
\end{equation}

\paragraph{Location Span Predictor}
\label{sec:loc-span-pred}
To predict the location span of given participants, the model calculates the probability distribution of the start and ending word of the location span. 

The span predictor takes the participant representation $H_{p^{t}}$, contextual representation $\widetilde{H}^D_t = [\widetilde{h}^1_t,\widetilde{h}^2_t..., \widetilde{h}^{|D|}_t]$ of each candidate word in paragraph $D$, and the probability distribution of start word $p^l_{t-1} \in \mathbb{R}^{|D|}$ predicted from last time step $t-1$ as input. 
To better utilize the symbolic model for location prediction, we use {location mask} to filter potential location span by two predefined rules:
(1) the location span is most likely relevant to an entity extracted in the graph;
(2) most location spans contain only nouns and adjectives. 
We observe that $98\%$ of the instances accord with the mentioned rules.

We reuse some of the operations from ProGlobal, and calculate the probability distribution at $t$-th time step (i.e., $p_t$) by the following formula, where $p_t^i$ represents the start probability distribution of location $i$ at the time step $t$.
\begin{equation}
\widetilde{H}^*_t = \sum_{i}^{|D|} p^i_{t-1} \cdot \widetilde{h}^i_t
\end{equation}
\begin{equation}
\varphi^i_t = LSTM([\widetilde{h}^i_t ; \widetilde{H}^*_t ;{H_{p^{t}}}])
\end{equation}
\begin{equation}
p_t = softmax(W_s\cdot[\widetilde{H}^D_t : \varphi^D_t]+b_s)
\end{equation}
We use the similar way to predict the probability of the ending word of the location span.

\subsection{Training and Inference}
The contextual representation model, graph network and prediction model are trained in an end-to-end manner. 
Model parameters are trained by minimizing the sum of the negative log likelihood calculated for the state change classification, location state classification and location span prediction. 
\par To better model the consistency between state predictor and location predictor, the model only infer the location span when the location state is classified into ``\textit{known location}". Otherwise, if the location state is ``\textit{does not exist}" or ``\textit{location unknown}", the location will be assigned to ``null" or ``unk". 

\section{Experiment}
In this section, we describe experimental settings, model comparison, ablation study and quantitative analysis. We abbreviate our approach as ProGraph.

\subsection{Task 1: Document Level Evaluation}

We first evaluate the performance of our model on the document-level task, which is to answer the first four questions as mentioned in Section \ref{section:task}. 


%

\begin{table}[htb] 
  \begin{minipage}[b]{0.48\textwidth} 
    \centering 
\begin{tabular}{r|c|c|c}
		\hline
		\multicolumn{1}{l|}{} & Precision     & Recall        & F1             \\ \hline
		QRN                           & 55.50          & 31.30          & 40.00           \\
		EntNet                         & 50.20          & 33.50          & 40.20           \\
		Pro-Global                    & 46.70          & 52.40		   & 49.40           \\
		Pro-Struct                     & 74.20          & 42.10          & 53.75          \\
		KG-MRC             		   & 64.52         & 50.68         & 56.77 \\ 
		
		\hline
		ProComp		   & 64.80		   & 38.10		   & 48.00 \\
		ProGraph	   & 67.30		   & 55.80		   & $\bm{61.00}$ \\ \hline
	\end{tabular}
	\caption{Results on document-level task. Our approach is abbreviated as ProGraph. ProComp is the symbolic-based baseline and other systems are strong neural baselines.}
	\label{table:task1-result}

  \end{minipage}%
  \ \ \ \ \ \ \ 
  \begin{minipage}[b]{0.48\textwidth} 
    \centering
\begin{tabular}{l|ccc}
		\hline
		\multirow{2}{*}{Model}  & \multicolumn{3}{c}{Performance} \\
		\cline{2-4}
		&P    &R     &F1\\\hline
		ProGraph    &    67.30     &  55.80        &  61.00        \\
		\ \ \ -w/o location mask & 66.50 & 53.80 & 59.50\\
		\ \ \ -w/o attribute    &    63.10       & 55.90        &  59.30        \\
		\ \ \ -w/o entire graph          &   62.10        &  46.90        &  53.40       \\
		
		\hline
	\end{tabular}
	\caption{Ablation experiments on PROPARA. We eliminate the components of graph-based reasoning model and then we eliminate the whole graph-based learning model (entire graph).}
	\label{table:ablation-experiment}
  \end{minipage} 
\end{table}

		

Table \ref{table:task1-result} reports the results on document-level task.
We also report the performance of our re-implemented ProComp system, with some modifications as described previously.
We compare ProGraph with the pure symbolic-based system ProComp described before.
As shown in the table, our system achieves 17.70\% absolute improvements in Recall and 13\% improvement on the F1 compared with ProComp. 
Moreover, our system also outperforms previous strong neural baselines with 61.00\% F1 score.
Our model also outperforms KG-MRC, the most related work to ProGraph, with 4.23\% absolute performance gain in F1 score.
This observation indicates that integrating graph neural network with the symbolic system not only persists the in-domain reasoning ability of the symbolic system but also alleviates its shortcomings when learning the fluidity of concepts. 

To make further analysis about the effect of different components (i.e. variants of attributes, nodes or relations), we make ablation experiments. 
Table \ref{table:ablation-experiment} describes the result after eliminating different components of our model. 
We eliminate the graph-based representation learning model. Instead of predicting the state of the participant by its learned representation from the graph network, we replace it with a paragraph-based representation calculated by max-pooling over the contextual representation  $\widetilde{H}^D_t$. 
As shown in the table, eliminating the component of graph-based representation learning model from the overall model causes substantial performance drops
(61.0\% to 53.4\% on F1).
The result verifies that incorporating the graph network can enhance the system's ability to perform entity state reasoning.

Moreover, we eliminate the location mask, which is designed to shrink the number of candidate location words by using the extracted graph.
This operation leads to 0.8\%, 2.0\% and 1.5\% performance drops on precision, recall and F1, respectively. This result indicates that the symbolic model can provide a strong prior knowledge of the important information to the learning of the neural model.  

We further remove the attribute node from the graph, and find that this operation
causes 4.2\% and 1.7\% performance drops on precision and F1 respectively. This observation confirms that attributes inferred by the symbolic system are meaningful in guiding the learning of graph network. 

\begin{figure}[h]
	\centering
	\includegraphics[width=0.67\textwidth]{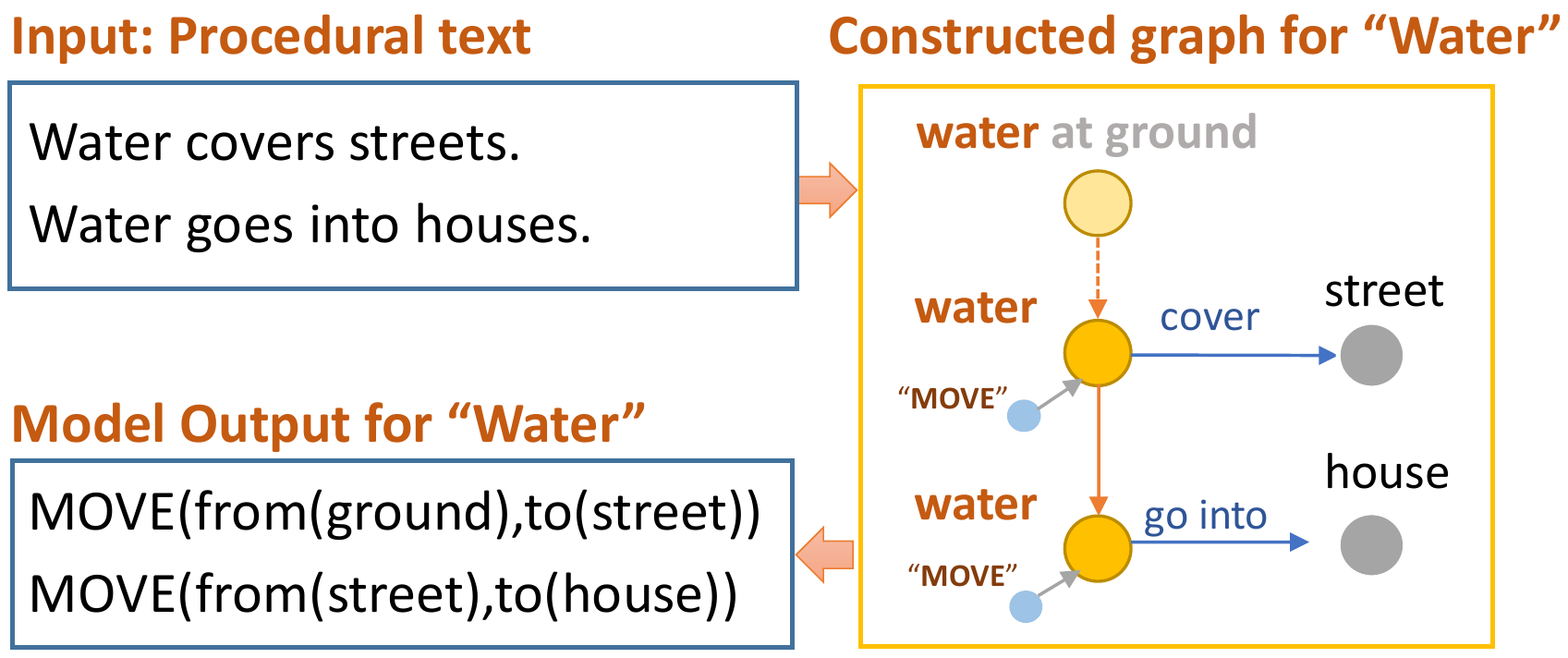}
	\caption{An example of model output. With given procedural text, the model first construct a graph for participant ``water". Then ProGraph perform reasoning over the graph and output the state change and the  position of ``water".}
	\label{fig:case-example}
\end{figure}

\paragraph{Case Study}
We provide a case study for the qualitative analysis of our model. 
As shown in Figure \ref{fig:case-example}, with two given sentences (i.e., ``Water covers streets. Water goes into houses."), our system can build the graph that captures the critical information from the sentences.
Afterwards, ProGraph learns the representation over the constructed graph and make a correct prediction with the prediction model. 
\subsection{Task 2: Sentence Level Evaluation}
The fine-grained sentence level task was proposed by Mishra et al. \cite{mishra2018tracking},
which requires the system to answer the last three questions as mentioned in Section\ref{section:task}.

\par As shown in Table \ref{table:task2-result}, our model outperforms previous strong baselines in all types of questions (4.9\%, 4.62\% and 3.55\% performance gain in Cat-1, Cat-2, Cat-3). 
This observation indicates that ProGraph is better at both location tracking and state change prediction because it can better model the correlation between participants and other entities. ProGraph also significantly outperforms ProComp in all types of question, which proves that integrating symbolic with the neural models enhances the reasoning ability.

\begin{table}[h]
	\centering
	\begin{tabular}{r|c|c|c|c|c}
		\hline
		\multicolumn{1}{l|}{} & \textbf{Cat-1} & \textbf{Cat-2} & \textbf{Cat-3} & \textbf{Macro-Avg} & \textbf{Micro-Avg} \\ \hline
		Human upper bound     & 91.67          & 87.66          & 62.96          & 80.76              & 79.69              \\ \hline
		EntNet \cite{henaffWSBL16}              & 51.62          & 18.83          & 7.77           & 26.07              & 25.96              \\
		QRN \cite{seo2016query}                & 52.37             & 15.51          & 10.92           & 26.26              & 26.49              \\
		Pro-Local \cite{mishra2018tracking}             & 62.65          & 30.50          & 10.35          & 34.50              & 33.96              \\
		Pro-Global \cite{mishra2018tracking}& 62.95 & 36.39          & 35.90          & 45.08              & 45.37              \\
		KG-MRC \cite{das2019kgmrc}         & 62.86          & 40.00			& 38.23 & 47.03     & 46.62     \\ 
		\hline
		ProComp  & 55.93 & 26.59 & 11.08 & 31.20 & 30.84 \\
		ProGraph & $\bm{67.76}$ & $\bm{44.62}$ & $\bm{41.78}$ & $\bm{51.39}$ & $\bm{51.53}$ \\\hline
	\end{tabular}
	\caption{Results on sentence-level task. Our approach is abbreviated as ProGraph. ProComp is the symbolic-based baseline and other systems are strong neural baselines.}
	\label{table:task2-result}
\end{table}

\subsection{Error Analysis and Discussion}
We analyze randomly selected 500 incorrectly predicted instances and summarize the major types of errors.
\par
The dominant type is that the model fails to distinguish the initial state of participants between ``\textit{does not exist}" or ``\textit{location unknown}" if the participant is not mentioned explicitly at the beginning of the process.
\par
The second type of errors is caused by failing to model consistency between the participant and its carrier. 
The state of the participant can be affected by the state of its carrier, but it is hard for the model to learn this consistency. 
For example, the procedural text mentions ``\textit{Soft tissues quickly decompose leaving behind hard bones or shells. Over time sediment builds over the top and hardens into rock.}". After this process, bones should be located in rock because sediment is the carrier of bones. However, the model fails to locate bones because it is not mentioned in the second sentence.
\par
The third type of errors is caused by the fact that the external knowledge resource (i.e., semantic lexicon) lack fine-grained knowledge about the meaning of verbs when associated with different entities. 
For instance, when the input is ``\textit{Rain clouds are stopped or slowed by mountains or wind.}", the model states that the clouds moves to the mountain while the answer is that they stay on the sky. The rule ``subject (stop by) object" will infer the state change ``subject (MOVE) to object" without considering different types of subjects. Therefore, the state of ``cloud" will be misclassified into ``MOVE" because it corresponds to the most situation when the action ``stop by" happened. 
\par
Moreover, the model may fail to identify real location through pronoun. For example, when the paragraph mentions ``\textit{The dead plants sink to the bottom of the swamps. Many more dead plants sink in the same area.}", the model outputs that the location of plants is ``\textit{same area}" while the golden answer is ``\textit{bottom of the swamps}". These two locations are the same with the referential relationship but the model fails to identify that. One intuition of solving this type of errors is to employ the co-reference resolution toolkits.





\section{Conclusion}
We present an approach, namely ProGraph, to improve entity state reasoning by integrating various types of knowledge into  neural network. 
We contribute by introducing
a graph-based reasoning framework, in which the graph construction process leverages factual, temporal and logical knowledge. The representations of nodes and the compositionality over a graph are modeled via neural models.
Results show that integrating the neural model and the symbolic model with the graph network significantly improves performance. Our model performs better than strong baselines on PROPARA dataset.

We suggest following directions for further research:

\begin{enumerate}
	\item Identifying ways to improve the construction of the participant-specific structure in a dynamic environment.
	\item Developing a better knowledge-enhanced model to automatically retrieve the external/world knowledge and integrate the knowledge for making the prediction.
	\item Handling the error at the initial state with prior/commonsense knowledge. 
\end{enumerate}

\bibliographystyle{acl}
\bibliography{coling2020}

\begin{thebibliography}{}

\bibitem[\protect\citename{Battaglia \bgroup et al.\egroup
  }2018]{battaglia2018}
Peter~W. Battaglia, Jessica~B. Hamrick, Victor Bapst, Alvaro
  Sanchez{-}Gonzalez, Vin{\'{\i}}cius~Flores Zambaldi, Mateusz Malinowski,
  Andrea Tacchetti, David Raposo, Adam Santoro, Ryan Faulkner, {\c{C}}aglar
  G{\"{u}}l{\c{c}}ehre, Francis Song, Andrew~J. Ballard, Justin Gilmer,
  George~E. Dahl, Ashish Vaswani, Kelsey Allen, Charles Nash, Victoria
  Langston, Chris Dyer, Nicolas Heess, Daan Wierstra, Pushmeet Kohli, Matthew
  Botvinick, Oriol Vinyals, Yujia Li, and Razvan Pascanu.
\newblock 2018.
\newblock Relational inductive biases, deep learning, and graph networks.
\newblock {\em CoRR}, abs/1806.01261.

\bibitem[\protect\citename{Bosselut \bgroup et al.\egroup }2017]{bosselut2018}
Antoine Bosselut, Omer Levy, Ari Holtzman, Corin Ennis, Dieter Fox, and Yejin
  Choi.
\newblock 2017.
\newblock Simulating action dynamics with neural process networks.
\newblock {\em CoRR}, abs/1711.05313.

\bibitem[\protect\citename{Clark \bgroup et al.\egroup }2018]{clark2018}
Peter Clark, Bhavana Dalvi, and Niket Tandon.
\newblock 2018.
\newblock What happened? leveraging verbnet to predict the effects of actions
  in procedural text.
\newblock {\em CoRR}, abs/1804.05435.

\bibitem[\protect\citename{Dalvi \bgroup et al.\egroup }2018]{dalviHTYC18}
Bhavana Dalvi, Lifu Huang, Niket Tandon, Wen{-}tau Yih, and Peter Clark.
\newblock 2018.
\newblock Tracking state changes in procedural text: a challenge dataset and
  models for process paragraph comprehension.
\newblock In {\em Proceedings of the 2018 Conference of the North American
  Chapter of the Association for Computational Linguistics: Human Language
  Technologies, {NAACL-HLT} 2018, New Orleans, Louisiana, USA, June 1-6, 2018,
  Volume 1 (Long Papers)}, pages 1595--1604.

\bibitem[\protect\citename{Das \bgroup et al.\egroup }2019]{das2019kgmrc}
Rajarshi Das, Tsendsuren Munkhdalai, Xingdi Yuan, Adam Trischler, and Andrew
  McCallum.
\newblock 2019.
\newblock Building dynamic knowledge graphs from text using machine reading
  comprehension.
\newblock {\em ICLR}.

\bibitem[\protect\citename{De~Cao \bgroup et al.\egroup }2018]{de2018question}
Nicola De~Cao, Wilker Aziz, and Ivan Titov.
\newblock 2018.
\newblock Question answering by reasoning across documents with graph
  convolutional networks.
\newblock {\em arXiv preprint arXiv:1808.09920}.

\bibitem[\protect\citename{Devlin \bgroup et al.\egroup }2018]{devlin2018bert}
Jacob Devlin, Ming-Wei Chang, Kenton Lee, and Kristina Toutanova.
\newblock 2018.
\newblock Bert: Pre-training of deep bidirectional transformers for language
  understanding.
\newblock {\em arXiv preprint arXiv:1810.04805}.

\bibitem[\protect\citename{Gori \bgroup et al.\egroup }2005]{gori2005new}
Marco Gori, Gabriele Monfardini, and Franco Scarselli.
\newblock 2005.
\newblock A new model for learning in graph domains.
\newblock In {\em Proceedings. 2005 IEEE International Joint Conference on
  Neural Networks, 2005.}, volume~2, pages 729--734. IEEE.

\bibitem[\protect\citename{Gupta and Durrett}2019]{gupta2019tracking}
Aditya Gupta and Greg Durrett.
\newblock 2019.
\newblock Tracking discrete and continuous entity state for process
  understanding.
\newblock {\em arXiv preprint arXiv:1904.03518}.

\bibitem[\protect\citename{Henaff \bgroup et al.\egroup }2016]{henaffWSBL16}
Mikael Henaff, Jason Weston, Arthur Szlam, Antoine Bordes, and Yann LeCun.
\newblock 2016.
\newblock Tracking the world state with recurrent entity networks.
\newblock {\em CoRR}, abs/1612.03969.

\bibitem[\protect\citename{Li \bgroup et al.\egroup }2015]{li2015gated}
Yujia Li, Daniel Tarlow, Marc Brockschmidt, and Richard Zemel.
\newblock 2015.
\newblock Gated graph sequence neural networks.
\newblock {\em arXiv preprint arXiv:1511.05493}.

\bibitem[\protect\citename{Manning \bgroup et al.\egroup
  }2014]{manning2014stanford}
Christopher Manning, Mihai Surdeanu, John Bauer, Jenny Finkel, Steven Bethard,
  and David McClosky.
\newblock 2014.
\newblock The stanford corenlp natural language processing toolkit.
\newblock In {\em Proceedings of 52nd annual meeting of the association for
  computational linguistics: system demonstrations}, pages 55--60.

\bibitem[\protect\citename{Mishra \bgroup et al.\egroup
  }2018]{mishra2018tracking}
Bhavana~Dalvi Mishra, Lifu Huang, Niket Tandon, Wen-tau Yih, and Peter Clark.
\newblock 2018.
\newblock Tracking state changes in procedural text: A challenge dataset and
  models for process paragraph comprehension.
\newblock {\em arXiv preprint arXiv:1805.06975}.

\bibitem[\protect\citename{Mitchell and Lapata}2010]{mitchell2010composition}
Jeff Mitchell and Mirella Lapata.
\newblock 2010.
\newblock Composition in distributional models of semantics.
\newblock {\em Cognitive science}, 34(8):1388--1429.

\bibitem[\protect\citename{Russell and Norvig}2002]{russell2002artificial}
Stuart Russell and Peter Norvig.
\newblock 2002.
\newblock Artificial intelligence: a modern approach.

\bibitem[\protect\citename{Scarselli \bgroup et al.\egroup
  }2009]{scarselliGTHM09}
Franco Scarselli, Marco Gori, Ah~Chung Tsoi, Markus Hagenbuchner, and Gabriele
  Monfardini.
\newblock 2009.
\newblock The graph neural network model.
\newblock {\em {IEEE} Trans. Neural Networks}, 20(1):61--80.

\bibitem[\protect\citename{Schuler}2005]{schuler2005verbnet}
Karin~Kipper Schuler.
\newblock 2005.
\newblock Verbnet: A broad-coverage, comprehensive verb lexicon.

\bibitem[\protect\citename{Seo \bgroup et al.\egroup }2016]{seo2016query}
Minjoon Seo, Sewon Min, Ali Farhadi, and Hannaneh Hajishirzi.
\newblock 2016.
\newblock Query-reduction networks for question answering.
\newblock {\em arXiv preprint arXiv:1606.04582}.

\bibitem[\protect\citename{Shi and Lin}2019]{shi2019simple}
Peng Shi and Jimmy Lin.
\newblock 2019.
\newblock Simple bert models for relation extraction and semantic role
  labeling.
\newblock {\em arXiv preprint arXiv:1904.05255}.

\bibitem[\protect\citename{Song \bgroup et al.\egroup }2018]{song2018exploring}
Linfeng Song, Zhiguo Wang, Mo~Yu, Yue Zhang, Radu Florian, and Daniel Gildea.
\newblock 2018.
\newblock Exploring graph-structured passage representation for multi-hop
  reading comprehension with graph neural networks.
\newblock {\em arXiv preprint arXiv:1809.02040}.

\bibitem[\protect\citename{Stanovsky \bgroup et al.\egroup
  }2018]{Stanovsky2018SupervisedOI}
Gabriel Stanovsky, Julian Michael, Luke~S. Zettlemoyer, and Ido Dagan.
\newblock 2018.
\newblock Supervised open information extraction.
\newblock In {\em NAACL-HLT}.

\bibitem[\protect\citename{Tandon \bgroup et al.\egroup }2018]{tandonDGYBC18}
Niket Tandon, Bhavana Dalvi, Joel Grus, Wen{-}tau Yih, Antoine Bosselut, and
  Peter Clark.
\newblock 2018.
\newblock Reasoning about actions and state changes by injecting commonsense
  knowledge.
\newblock In {\em Proceedings of the 2018 Conference on Empirical Methods in
  Natural Language Processing, Brussels, Belgium, October 31 - November 4,
  2018}, pages 57--66.

\bibitem[\protect\citename{Veli{\v{c}}kovi{\'c} \bgroup et al.\egroup
  }2017]{velivckovic2017graph}
Petar Veli{\v{c}}kovi{\'c}, Guillem Cucurull, Arantxa Casanova, Adriana Romero,
  Pietro Lio, and Yoshua Bengio.
\newblock 2017.
\newblock Graph attention networks.
\newblock {\em arXiv preprint arXiv:1710.10903}.

\end{thebibliography}

\end{document}